# Comparative Analysis of ChatGPT, GPT-4, and Microsoft Bing Chatbots for GRE Test

**Mohammad Abu-Haifa**
Forensics and Investigations, Thornton Tomasetti, Inc.
Orlando, Florida, USA
https://orcid.org/0009-0009-3554-8774

**Bara'a Etawi**
Computer Engineering Department, Yarmouk University
Irbid, Jordan

**Huthaifa Alkhatatbeh**
Ministry of Municipal Affairs
Irbid, Jordan
https://orcid.org/0000-0002-3017-1834

**Ayman Ababneh**
Civil Engineering Department, Jordan University of Science and Technology
Irbid, Jordan
https://orcid.org/0000-0002-6962-7368

**Corrospending Author:** Mohammad Abu-Haifa,
Email: MAbuHaifa@ThorntonTomasetti.com

**Abstract.** This research paper presents an analysis of how well three artificial intelligence chatbots: Bing, ChatGPT, and GPT-4, perform when answering questions from standardized tests. The Graduate Record Examination is used in this paper as a case study. A total of 137 questions with different forms of quantitative reasoning and 157 questions with verbal categories were used to assess their capabilities. This paper presents the performance of each chatbot across various skills and styles tested in the exam. The proficiency of these chatbots in addressing image-based questions is also explored, and the uncertainty level of each chatbot is illustrated. The results show varying degrees of success across the chatbots, where GPT-4 served as the most proficient, especially in complex language understanding tasks and image-based questions. Results highlight the ability of these chatbots to pass the GRE with a high score, which encourages the use of these chatbots in test preparation. The results also show how important it is to ensure that, if the test is administered online, as it was during COVID, the test taker is segregated from these resources for a fair competition on higher education opportunities.

**Keywords:** Technology-enhanced learning; Graduate record examination; ChatGPT; GPT-4; Microsoft Bing

# 1. Introduction

## 1.1 Graduate Record Examination

The Graduate Record Examinations (GRE) is a standardized test that is administrated by the Educational Testing Service (ETS, 2023a), commonly known as ETS, to assess certain skills of graduate school applicants (Moneta-Koehler et al., 2017). The GRE aims to measure the test-taker's readiness for graduate-level academic work and is often used by universities in the United States and certain other countries to evaluate applicants to their graduate programs in a wide range of subject areas, such as the sciences, humanities, and engineering (Bleske-Rechek & Browne, 2014). Therefore, GRE offers a common metric that educational institutions can adopt to compare applicants from different backgrounds.

The GRE consists of three main sections: verbal reasoning, quantitative reasoning, and analytical writing (Klieger et al., 2022). In the reasoning sections, there are four multiple-choice sessions: two for verbal and two for quantitative. Each session has a different number of questions: the first session for each reasoning section has 12 questions, and the second session has 15 questions (ETS, 2023a). Verbal reasoning section measures the candidate's vocabulary and reading comprehension, while quantitative reasoning measures problem-solving ability in four major skills: arithmetic, algebra, geometry, and data analysis (Liu et al., 2015). The analytical writing section comprises two tasks, one that requires the student to analyze an argument and another in which they need to develop an essay stating their perspective on a given issue. This section evaluates the student's ability to write a coherent and effective essay (Roohr et al., 2022).

## 1.2 Natural Language Processing Chatbots

Artificial intelligence (AI) tools refer to a collection of software applications that aim to automate routine tasks, analyze data sets, and provide data-driven insights to make decisions by identifying the patterns in processed datasets (M. Y. Ali et al., 2020; Phillips-Wren, 2012). One of the recent applications of AI tools is natural language processing (NLP), which focuses on the interaction between computers and humans using natural language, specifically the ability of computers to understand, interpret, and generate human language (Hirschberg & Manning, 2015). NLP tools can process large amounts of text data and identify patterns that would be challenging for humans to uncover (Chowdhary & Chowdhary, 2020). NLP has been an active area of research for several years, but its current swift advancements reflect a transformation that was initially gradual but has now experienced rapid acceleration (Rudolph et al., 2023).

AI chatbots, such as ChatGPT (OpenAI, 2023a), Microsoft Bing chatbot (OpenAI, 2023a), GPT-4 (OpenAI, 2023b), and Bard (Google, 2023), use NLP to understand and generate human-like responses in conversations using voice or text. These chatbots operate within different contexts, providing a range of services and support, which helped them to gain popularity in recent years. While these chatbots may generate similar results, there are notable differences between them in terms of integrated models, enabled features and plugins, response time, and accuracy of the response.

Recently, several studies have evaluated the efficiency of NLP chatbots in public and mental health (Biswas, 2023b), higher education (Crompton & Burke, 2023), global warming (Biswas, 2023a), solving programming bugs (Surameery & Shakor, 2023), writing literature reviews (Haman & Školník, 2023), promoting learning and teaching (Baidoo-Anu & Ansah, 2023; Ouyang et al., 2023), writing patient clinic letters (S. R. Ali et al., 2023), and improving various research methods (Burger et al., 2023). The performance of using these chatbots in solving and passing some exams was also investigated in the literature, such as the medical USMLE exam (Gilson et al., 2023) and law school exams (Choi et al., 2023; Hargreaves, 2023). For instance, Kung et al (2023) showed that ChatGPT performed near the passing threshold for USMLE (Steps 1, 2, and 3) without any specialized training or reinforcement. In addition, the capability of the chatbots was investigated in more educational aspects like summarizing and extracting specific information from the text (Fang et al., 2015), building a dialogue system that can assist student's knowledge (Abro et al., 2022), analyzing medical sentences (Dominy et al., 2022), scoring essay questions and detecting plagiarism (Khurana et al., 2023). However, the significant capability of NLP chatbots has raised concerns about the use of these chatbots as a tool for academic misconduct in online exams (Susnjak, 2022). Also, most of the studies mentioned above are limited to examine the potential of OpenAI GPT models.

The primary aim of this study is to evaluate the independent capability of different NLP chatbots in passing a standardized test like the GRE. Specifically, this study focuses ChatGPT, GPT-4, and Microsoft Bing performance in verbal and quantitative reasoning sections of the GRE. This research provides insights into AI technology advancements and its potential for effective technology-assisted learning strategies in education. Additionally, this study identifies the areas where chatbots may struggle. This can help in personalized learning, test preparation, and can guide institutions in optimizing their teaching methodologies.

## 2. Methodology

### 2.1 Input Questions

This study adopts 331 reasoning questions (i.e., verbal and quantitative) that are obtained from the official ETS website (ETS, 2023a). The questions are divided as follows: 174 quantitative reasoning questions and 157 verbal reasoning questions obtained from (ETS, 2023d, 2023c, 2023b). All dataset questions were screened, and quantitative reasoning questions containing visual images such as plots or geometries were removed from the main dataset. These image-based questions were kept at a separate dataset, and each image was uploaded on Google to evaluate the capability of the chatbots to read the image from the provided Google link and answer the given question. The main dataset after removing the questions containing images consists of 294 questions (137 quantitative and 157 verbal). This means that 37 image-based quantitative questions were kept in the second dataset. The two datasets were advanced to the encoding as shown in the following section.

The quantitative reasoning questions examine four major skills: arithmetic, algebra, geometry, and data analysis, and they encompass four question styles. These include three multiple-choice styles (quantitative comparison, only one answer is correct, and one or more answers are correct), as well as one numeric entry style that does not provide any possible answers. On the other hand, the verbal reasoning questions examine three primary skills: reading comprehension, text completion, and sentence equivalence. More than one blank can exist in each verbal question with several choices for each blank. The verbal reasoning questions, in the considered references, are grouped into several exams, each of which is categorized according to its difficulty level as easy, medium, hard, or mixed.

Table 1 provides a breakdown of quantitative questions in terms of the number of questions included in the study. In contrast, the distribution of verbal questions is presented in Table 2, considering the skill set to be tested and the level of difficulty associated with each question.

**Table 1: Distribution of quantitative questions (main dataset) that used in this study**

|  | Arithmetic | Algebra | Geometry | Data analysis | **Total** |
|---|---|---|---|---|---|
| Quantitative comparison | 23 | 14 | 7 | 10 | 54 |
| One answer is correct | 12 | 13 | 8 | 14 | 47 |
| One or more answers are correct | 8 | 2 | 1 | 6 | 17 |
| Numeric entry | 9 | 6 | 0 | 4 | 19 |
| **Total** | 52 | 35 | 16 | 34 | **137** |

**Table 2: Distribution of verbal questions that used in this study**

|  | Reading comprehension | Text completion | Sentence equivalence | **Total** |
|---|---|---|---|---|
| Easy | 9 | 9 | 5 | 23 |
| Medium | 11 | 8 | 8 | 27 |
| Hard | 10 | 8 | 7 | 25 |
| Mixed | 41 | 24 | 17 | 82 |
| **Total** | 71 | 49 | 37 | **157** |

## 2.2 Encoding and Adjudication

The questions from both datasets were entered into the three chatbots (i.e., Bing, ChatGPT, and GPT-4) by their respective styles (i.e., multiple-choice, or numeric entry) and to test chatbot's stability in solving both verbal and quantitative questions. This testing consisted of two separate trials:

1. The first trial involved entering the question into the chatbot as is, along with all possible answers, if the style is multiple-choice, to verify its ability to answer correctly.
2. During the second trial, the stability and certainty of the chatbot were subjected to further examination. To assess its reliability, the chatbot was prompted to re-evaluate its initial response by encoding a statement asking for confirmation regarding its level of confidence in delivering the correct answer. The following statement was encoded for this purpose: "Are you sufficiently confident with your answer? Please conduct a reanalysis to ensure accuracy and subsequently provide the answer once again.".

To minimize the risk of systematic errors introduced using inflexible wording, the encoders purposefully varied the lead-in prompts. Additionally, in order to minimize any potential bias due to memory retention, a fresh chat session was initiated for each individual entry in the chatbot.

When there is the potential for multiple correct answers for quantitative questions or when verbal questions feature more than one blank to fill, the chatbot's answer is judged correct only if every choice and blank were correctly answered. Similarly, for numeric entry questions, an answer is considered correct if the answer provided is exact or its equivalent. For example, an answer of 0.4 would be equivalent to 4/10 for the encoder and would be considered correct.

## 3. Analysis and Results

### 3.1 Quantitative Reasoning Questions

#### 3.1.1 Main Dataset
As previously mentioned, quantitative questions have been segregated into two primary datasets: the main dataset, which contains 137 questions, and a secondary image-based dataset with 37 questions. Following encoding of all questions, the results suggest that the chatbots performances are promising. However, Bing chatbot struggles to provide accurate solutions for more difficult problems, particularly those in the data analysis and algebra category. Overall, Bing provided accurate solutions to 67 out of 137 quantitative reasoning questions, indicating a success rate of 48.9%. Bing's performance on the main dataset could be attributed to the complexity of such mathematical problems, which may require extensive contextual analysis, a deeper understanding of mathematical concepts, and algorithms that Bing may not have fully acquired. ChatGPT and GPT-4 showed superior performance in the quantitative reasoning questions (main dataset) and provided accurate solutions to 79 and 114 out of 137, indicating a success rate of 57.66% and 83.21%, respectively.

Figure 1 presents details regarding the percentage of correct answers in each skill provided by the chatbots. In detail, Bing's accuracy in providing correct answers for data analysis and algebra questions was observed to be less than 50%. The performance of Bing in arithmetic and geometry questions was modestly better.

Results show that Bing's performance is satisfactory when no possible answers are provided (i.e., numeric entry style). Its performance in 'one answer is correct' questions is noted to be better than in 'quantitative comparison' and 'one or more answers are correct' questions, as illustrated in Figure 2 This means that Bing becomes more confused once it is asked to select more than one answer or to compare between two quantities. The performance of ChatGPT in providing correct answers for algebra questions was observed to be the least, indicating that it struggles with algebra in comparison with the remaining skills. Similarly, GPT-4 also faced difficulties with algebra, making it the most challenging skill for both models. However, GPT-4's overall performance was deemed satisfactory, as it provided correct answers for all skills with a minimum success rate of 74.29%. One notable observation is that ChatGPT faced challenges when multiple answers are correct (i.e., one or more correct answers style), leading to higher mathematical uncertainty in its response selection with a success rate of 23.53%. Conversely, both ChatGPT and GPT-4 performed well in the numeric entry style. It is worth noting that GPT-4 accurately answered all numeric entry questions.

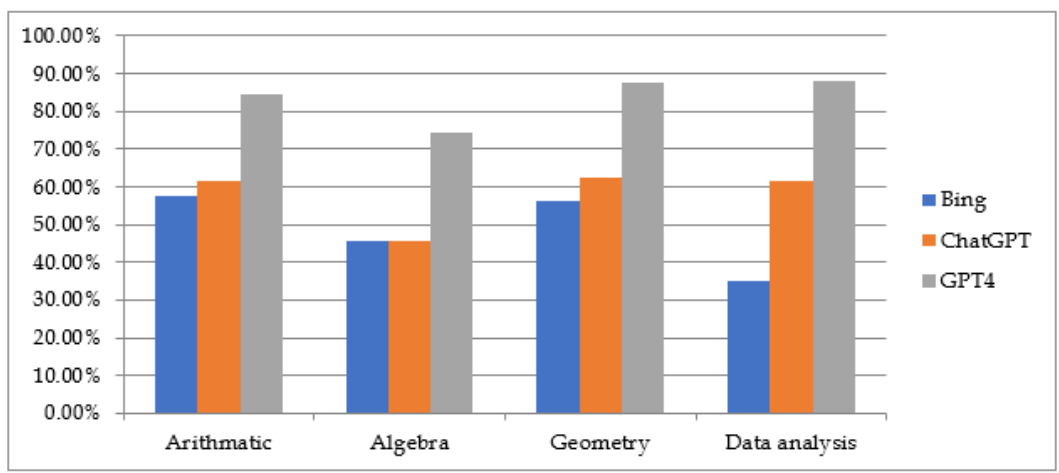

**Figure 1: Percent of correct answers for quantitative questions in each skill**

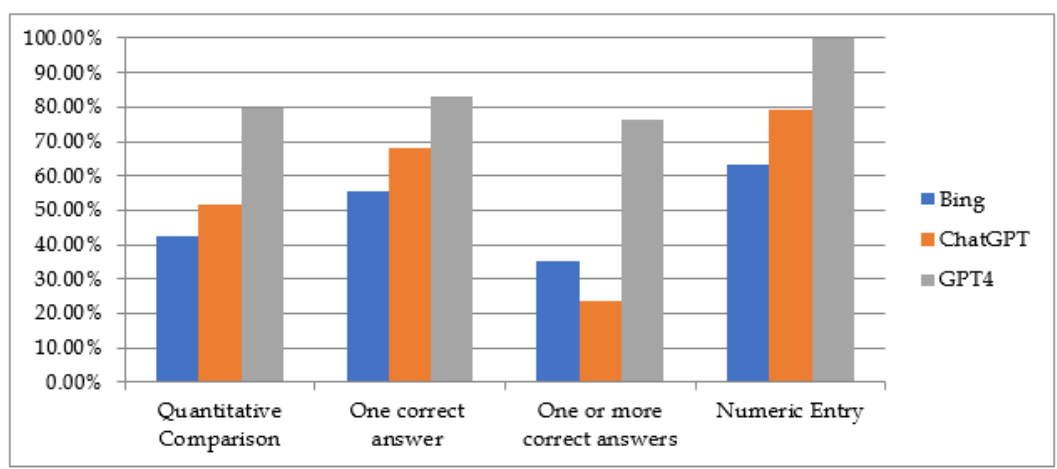

**Figure 2: Percent of correct answers for quantitative questions in each style**

Table 3 presents details regarding the performance of all chatbots across various skills and styles for the main dataset. It displays success rates expressed as fractions, where the numerator represents the number of correct answers, and the denominator represents the total number of questions within each skill or style. Results indicate that the three chatbots exhibit better performance in the absence of possible answers. Therefore, it is advisable to use numeric entry style in quantitative questions if these chatbots are to be used in test preparation, thus better outcomes. The 'one or more answers are correct' style is featured in 17 of the questions within the main dataset, as illustrated in Table 1. All questions in this style had multiple correct answers. When responding to these questions, Bing chatbot correctly identified all correct answers in only six instances, while the remaining 11 questions contained at least one incorrect answer (judgment criteria). The analysis showed that for three of these 11 questions, Bing was not able to select any correct answers. Nonetheless, for the other questions, Bing identified at least one correct answer. In contrast, ChatGPT correctly identified all correct answers in only 4 instances. In the remaining 13 questions, it was observed that ChatGPT was unable to select any correct answers for four of these 13 questions. ChatGPT successfully identified at least one correct answer for the remaining questions. Conversely, GPT-4 accurately identified all correct answers in 13 instances out of the 17 questions. GPT-4 was also unable to select any correct answer in only one question and there was at least one correct answer for three questions. This means that GPT-4 significantly outperforms Bing and ChatGPT in solving the quantitative questions.

**Table 3: Success rates summary of the chatbots for the main dataset**

| Skill | Chatbot | Quantitative comparison | One correct answer | One or more correct answers | Numeric entry | **Sum** |
|---|---|---|---|---|---|---|
| Arithmetic | Bing | 14/23 | 7/12 | 2/8 | 7/9 | **30/52** |
| | ChatGPT | 14/23 | 8/12 | 2/8 | 8/9 | **32/52** |
| | GPT-4 | 19/23 | 10/12 | 6/8 | 9/9 | **44/52** |
| Algebra | Bing | 5/14 | 9/13 | 0/2 | 2/6 | **16/35** |
| | ChatGPT | 4/14 | 8/13 | 1/2 | 3/6 | **16/35** |
| | GPT-4 | 11/14 | 8/13 | 1/2 | 6/6 | **26/35** |
| Geometry | Bing | 3/7 | 6/8 | 0/1 | 0/0 | **9/16** |
| | ChatGPT | 3/7 | 7/8 | 0/1 | 0/0 | **10/16** |
| | GPT-4 | 6/7 | 8/8 | 0/1 | 0/0 | **14/16** |
| Data analysis | Bing | 1/10 | 4/14 | 4/6 | 3/4 | **12/34** |
| | ChatGPT | 7/10 | 9/14 | 1/6 | 4/4 | **21/34** |
| | GPT-4 | 7/10 | 13/14 | 6/6 | 4/4 | **30/34** |
| **Sum** | **Bing** | **23/54** | **26/47** | **6/17** | **12/19** | **67/137** |
| | **ChatGPT** | **28/54** | **32/47** | **4/17** | **15/19** | **97/137** |
| | **GPT-4** | **43/54** | **39/47** | **13/17** | **19/19** | **114/137** |

The 'one or more answers are correct' style is featured in 17 of the questions within the main dataset, as illustrated in Table 1. All questions in this style had multiple correct answers. When responding to these questions, Bing chatbot correctly identified all correct answers in six instances, while the remaining 11 questions contained at least one incorrect answer (judgment criteria). The analysis showed that for three of these 11 questions, Bing was not able to select any correct answers. Nonetheless, for the other questions, Bing identified at least one correct answer. In contrast, ChatGPT correctly identified all correct answers in only four instances. In the remaining 13 questions, it was observed that ChatGPT was unable to select any correct answers for four of these 13 questions. ChatGPT successfully identified at least one correct answer for the remaining questions. Conversely, GPT-4 accurately identified all correct answers in 13 instances out of the 17 questions. GPT-4 was also unable to select any correct answer in only one question and there was at least one correct answer for three questions. This means that GPT-4 significantly outperforms Bing and ChatGPT in solving the quantitative questions.

Upon requesting chatbots to redo the analysis and provide answers for the same set of questions in a second trial, we observed that Bing's performance remained stable. Remarkably, Bing didn't change any correct answers to incorrect ones. However, out of the 70 questions that were previously answered incorrectly, only three answers were modified. While the answers were changed in these three questions, they remained incorrect. Thus, Bing's success rate and overall performance in quantitative questions remained consistent. On the other hand, neither ChatGPT nor GPT-4 changed any incorrect answer to a correct one in the second trial. However, ChatGPT changed the answers of three questions that were initially answered correctly to incorrect ones. These questions pertained to arithmetic skills, with one being in numeric entry style and the other two in quantitative comparison style. This observation suggests that ChatGPT exhibits uncertainty to some extent when dealing with arithmetic questions. Likewise, GPT-4 also modified the correct answers of two questions, leading to a change to incorrect. Among these questions, one falls under the category of arithmetic skills, while the other belongs to the data analysis style.

### 3.1.2 Image-based dataset

As previously mentioned, the secondary dataset consists of 37 image-based quantitative questions. Each question's image was uploaded on Google, and a corresponding Google link was generated. The questions were input into the chatbots as they were originally conveyed, followed by the statement: "You can locate the image by visiting the following link: [Google link].", along with all possible answers. Notably, the recent update to GPT-4 now permits direct image uploads. Consequently, this paper also assesses GPT-4's capability to address questions upon the direct upload of images. The chatbot responses on this dataset can be categorized into five main groups: i) correct analysis leading to a correct final answer, ii) incorrect analysis leading to an incorrect final answer, iii) correct analysis resulting in an incorrect final answer, iv) incorrect analysis but still yielding a correct final answer, and v) inability of the chatbot to interpret the image and provide any answer due to the nature of the image, link, or the complexity of the question itself. An example of ChatGPT's response, indicating its inability to solve the question and provide an answer, is illustrated in Figure 3 In this paper and due to the predominance of multiple-choice questions in the GRE exam, the success rate of the chatbot in this dataset was calculated based on the final answer. This calculation involved considering the combined outcomes of groups (i) and (iv).

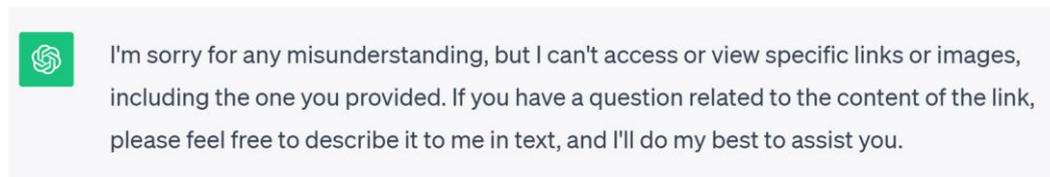

I'm sorry for any misunderstanding, but I can't access or view specific links or images, including the one you provided. If you have a question related to the content of the link, please feel free to describe it to me in text, and I'll do my best to assist you.

**Figure 3: ChatGPT's response to queries requiring tasks beyond its capabilities**

Until the date of writing this paper, the analysis of the second image-based dataset indicates that Bing chatbot generally struggles to extract the information from the images to provide accurate solutions. Bing was unable to read the images and provide any answer for 12 questions. Out of the remaining 25 questions, it provided incorrect solutions for 15 questions, but managed to produce the correct final answers for four of these questions. Seven of the remaining 10 questions were analyzed correctly with a correct final answer, while the other question resulted in an incorrect answer despite the correct analysis. It exhibited inappropriate utilization of the text-based information in instances where answer choices were available. This led to inaccurate solutions for most of the questions, except for the five cases where the final answer happened to be correct despite faulty analysis of four of those questions.

ChatGPT was unable to interpret the image and provide answers for the majority of image-based questions. In response to 19 questions, it indicated its inability to access the image through the provided link, as presented in Figure 3. For some of these questions, the chatbot resorted to providing general advice related to the topic based on the keywords mentioned in the question. This advice may not have been directly relevant or specific to the image in question. At times, this guidance led to a general procedure to find the answer (Figure 4a), while in other cases, it provided a direct final answer (Figure 4b). In the remaining 18 questions, ChatGPT relied solely on the text-based information provided within the question to attempt an answer. It provided incorrect calculations for the solutions in 15 questions. However, it managed to give the correct final answers for two of these questions, despite the overall flawed analysis in those cases. Nevertheless, two question was solved correctly in terms of both the analysis and the final answer. The calculations for the remaining question were accurate; however, the final answer provided was incorrect. Consequently, based on this evaluation, the success rate of ChatGPT in accurately addressing image-based questions was observed to be 4 out of the 37 questions posed.

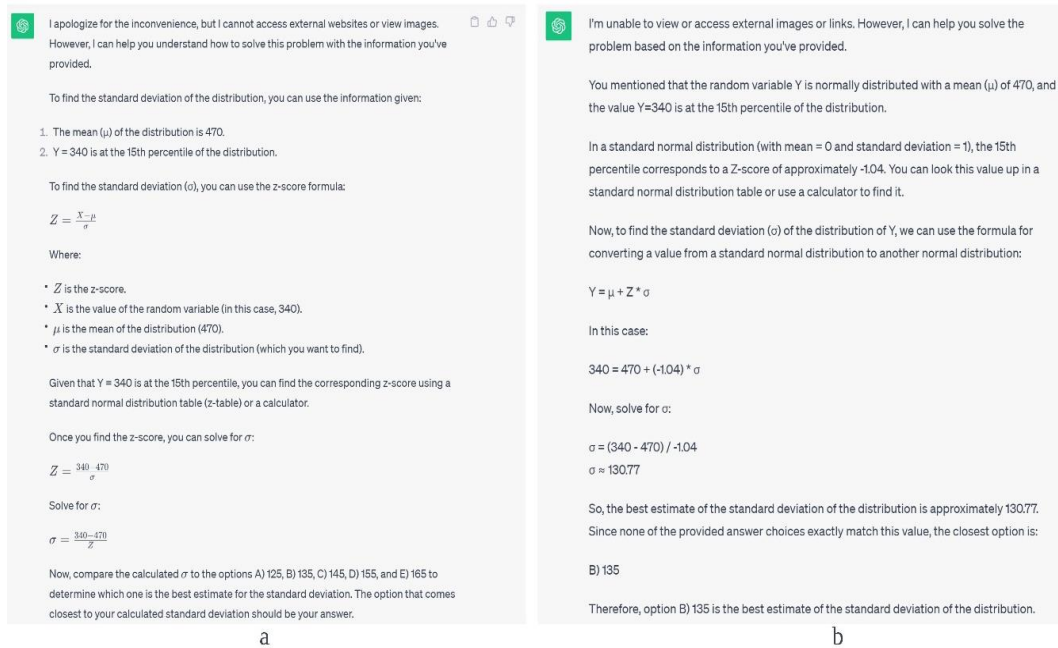

a                                         b

**Figure 4: Examples of ChatGPT responses for image-based questions with; a) general procedure and b) incorrect analysis with a direct final answer**

When the image was provided via external link, GPT-4 performed better in terms of image-based questions as it apologized to do an analysis for 11 questions only. It provided the correct final answers for 16 questions, and in 15 of these questions, the analysis was accurate. GPT-4 incorrectly analyzed seven of the questions, resulting in incorrect final answers, and there were three questions where the chatbot analyzed the images correctly but still provided incorrect final answers. This indicates a higher capability of GPT-4 to read and interpret information from images, along with the text-based information presented in the questions. On the other hand, the direct upload of the image to GPT-4 resulted in slightly different performance. It only apologized for its inability to analyze four questions. For five questions, it provided a general procedure to solve them without delivering a final answer. Furthermore, it supplied the correct final answers for 12 questions, with 10 of these accompanied by accurate analyses. However, GPT-4 erred in analyzing 13 questions, leading to incorrect final answers. Additionally, there were three questions in which the chatbot correctly analyzed the images but still provided incorrect final answers. Nevertheless, since this feature is specific to GPT-4, the results of the direct upload were not considered when comparing the chatbots in this paper. Instead, the comparison focused on the ability of all chatbots to analyze images from external links.

Table 4 presents a detailed summary for the performance of the chatbots on the image-based dataset. Findings confirm that Bing and ChatGPT are more likely to offer an inaccurate solution, leading to an incorrect final answer, even when they can interpret the image. In contrast, GPT-4 exhibits the potential to deliver both an accurate solution and a correct final answer. Furthermore, ChatGPT encounters difficulties when handling image-based questions, as its reliance on the text-based information provided within the question can lead to inaccuracies in the solutions it provides. It is worth noting that the quality of some images used in the dataset was not perfect. This means that certain fonts and words in the provided images were unclear, which could hinder the chatbot's ability to properly analyze the question. If the image resolution is improved and the information within it is clear, these chatbots could achieve higher scores than what was observed in this paper. However, it is advisable for students to be attentive when addressing image-based questions using these chatbots for test preparation.

**Table 4: Summary of chatbots results of the image-based dataset**

| Chatbot | Group (i)[1] | Group (ii)[2] | Group (iii)[3] | Group (iv)[4] | Group (v)[5] | Succes rate |
|---------|--------------|---------------|----------------|---------------|--------------|-------------|
| Bing | 7 | 11 | 3 | 4 | 12 | 11/37 |
| ChatGPT | 2 | 13 | 1 | 2 | 19 | 4/37 |
| GPT-4 | 15 | 7 | 3 | 1 | 11 | 16/37 |

[1] A correct analysis with a correct final answer
[2] An incorrect analysis with an incorrect final answer
[3] A correct analysis with an incorrect final answer
[4] An incorrect analysis with a correct final answer
[5] No access to the provided image

Upon requesting chatbots to redo the analysis and provide answers for the same set of questions in a second trial, we observed that Bing's corrected only the analysis of one question that was inaccurately analyzed from Group (ii) but still generated inaccurate final answers in Group (iii), resulting in the same success rate of 11 out of 37 questions. Furthermore, both ChatGPT and GPT-4 maintained the correctness of their analyses, ensuring that no correct analysis turned incorrect. However, GPT-4 changed the final correct answer for one question in Group (i), despite reiterating the correct analysis for the entire 15 questions that were solved accurately in the first trail. As for ChatGPT, it modified the analysis of five questions, all of which initially had incorrect analysis and final answers. Among these, one question was solved correctly in terms of analysis and final answer, while the remaining four were analyzed correctly but with keeping the incorrect final answers. Consequently, ChatGPT achieved a success rate of five correct final answers out of 37 questions after the second trial, with two of them having incorrect analysis. On the other hand, GPT-4 attained a success rate of 15 correct final answers out of 37 questions after the second trial, even with maintaining the accuracy of its correct analysis throughout. A summary of the results from the second trial on the image-based dataset is provided in Table 5. Results indicate that despite an increase in ChatGPT's success rate after the second trial, where it revised the analysis of more questions, Bing and GPT-4 still demonstrate better certainty in these types of questions.

**Table 5: Summary of chatbots results for the second trial of image-based dataset**

| Chatbot | Group (i) | Group (ii) | Group (iii) | Group (iv) | Group (v) | Succes rate |
|---------|-----------|------------|-------------|------------|-----------|-------------|
| Bing | 7 | 10 | 4 | 4 | 12 | 11/37 |
| ChatGPT | 3 | 8 | 5 | 2 | 19 | 5/37 |
| GPT-4 | 14 | 7 | 4 | 1 | 11 | 15/37 |

### 3.2 Verbal Reasoning Questions

The results of evaluating the performance of the chatbots on a set of 157 verbal questions is presented in this section. As mentioned previously, the verbal questions consisted of three different skills: reading comprehension, text completion, and sentence equivalence. Additionally, the questions were categorized into different difficulty levels: easy, medium, hard, or mixed style. After encoding all the questions, the results highlighted that there are varying levels of accuracy and success rates among the chatbots in handling verbal questions, similar to what was observed with the quantitative questions. Notably, GPT-4 exhibited the highest proficiency among the chatbots when it came to answering verbal questions. However, the three chatbots have shown a relatively higher level of accuracy in handling the verbal questions in comparison with the quantitative questions.

Overall, Bing successfully answered 103 out of 157 questions, yielding a success rate of 65.61%. ChatGPT achieved a higher performance by accurately answering 112 out of 157 questions, resulting in a success rate of 71.34%. On the other hand, GPT-4 demonstrated the highest ability in answering the verbal questions, with a success rate of 87.26%. It provided correct responses to 137 out of the 157 questions. Figure 5 visualizes the percentage of correct answers achieved by the chatbots in each skill. Specifically, the analysis reveals that Bing has the lowest accuracy (59.15%) in providing correct answers for reading comprehension questions, while its performance in text completion and sentence equivalence questions is relatively better. GPT-4 also displays a slight difficulty in answering reading comprehension questions, although it achieves a higher success rate of 83.10% than other chatbots. Conversely, the results indicate that sentence equivalence skill presents more challenge for ChatGPT in comparison with other skills, as it obtained a success rate of 67.57%.

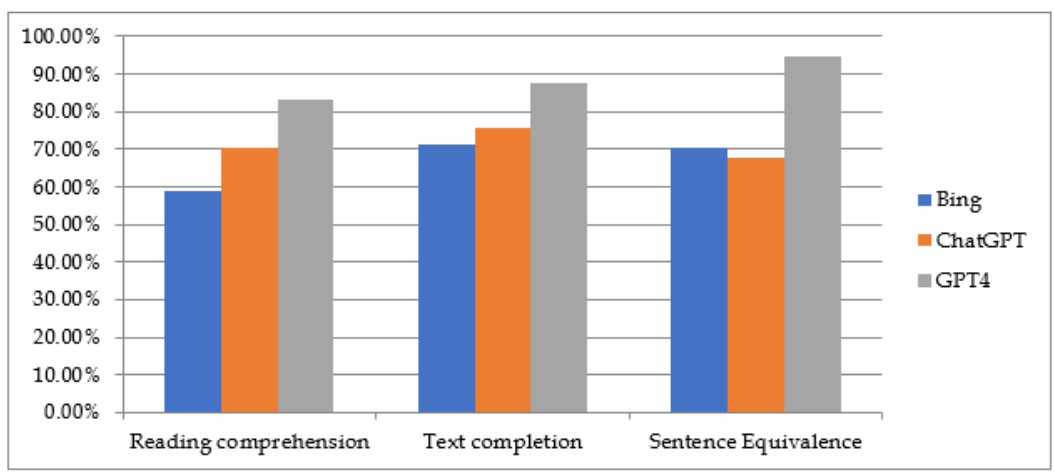

**Figure 5: Performance of all chatbots in answering verbal questions across several skills**

Figure 6 illustrates the performance of the chatbots in answering the verbal questions across various difficulty levels. Bing's performance was relatively better in medium questions compared to hard questions, while facing challenges in easy questions with the lowest success rate in this level (60.87%). For ChatGPT, the success rate demonstrates an increase with higher difficulty levels. In contrast, when the question becomes easier, GPT-4 exhibits a higher likelihood of providing correct answers. The varying performance of the chatbots across different difficulty levels can be attributed to the differences in their design, training data, and the way they process language. ChatGPT's performance in hard questions could be due to its extensive training on a diverse range of text data, which allows it to handle more complex language and reasoning. On the other hand, GPT-4's ability to perform better on easier questions might be because of its emphasis on common knowledge and factual information, making it excel in simpler, fact-based questions.

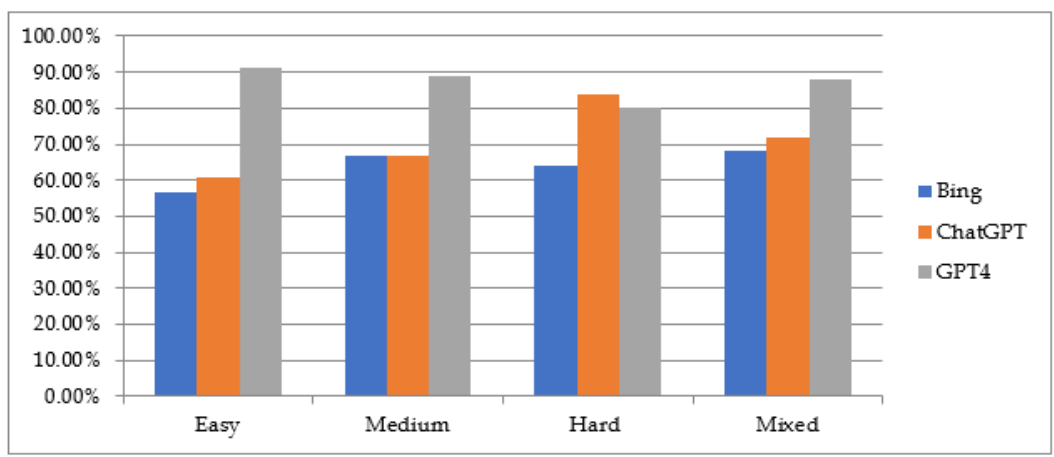

**Figure 6: Performance of chatbots in answering verbal questions across different difficulty levels**

A comprehensive overview of the performance of the three chatbots in addressing the three verbal skills, classified by difficulty levels and mixed styles, is presented in Table 6. For reading comprehension, the performance of ChatGPT and GPT-4 consistently surpassed that of Bing across all difficulty levels. GPT-4 managed to answer all the easy text completion questions correctly. It also consistently outperformed Bing and ChatGPT in all difficulty levels of sentence equivalence, achieving a perfect score in the medium and hard categories. Even while ChatGPT was still competent, its performance in this skill was the lowest when compared to GPT-4 and Bing. To sum up, in verbal questions, GPT-4's strength may lie in simplicity, ChatGPT's in complexity, and Bing's in a balanced approach that combines both.

**Table 6: Comparison of the chatbots' performance in verbal skills across several difficulty levels**

| Skill | Chatbot | Easy | Medium | Hard | Mixed | **Sum** |
|---|---|---|---|---|---|---|
| Reading comprehension | Bing | 4/9 | 7/11 | 6/10 | 25/41 | **42/71** |
| | ChatGPT | 4/9 | 7/11 | 9/10 | 30/41 | **50/71** |
| | GPT-4 | 8/9 | 10/11 | 7/10 | 34/41 | **59/71** |
| Text completion | Bing | 7/9 | 5/8 | 4/8 | 19/24 | **35/49** |
| | ChatGPT | 8/9 | 5/8 | 5/8 | 19/24 | **37/49** |
| | GPT-4 | 9/9 | 6/8 | 6/8 | 22/24 | **43/49** |
| Sentence equivalence | Bing | 2/5 | 6/8 | 6/7 | 12/17 | **26/37** |
| | ChatGPT | 2/5 | 6/8 | 7/7 | 10/17 | **25/37** |
| | GPT-4 | 4/5 | 8/8 | 7/7 | 16/17 | **35/37** |
| **Sum** | **Bing** | **13/23** | **18/27** | **16/25** | **56/82** | **103/157** |
| | **ChatGPT** | **14/23** | **18/27** | **21/25** | **59/82** | **112/157** |
| | **GPT-4** | **21/23** | **24/27** | **20/25** | **72/82** | **137/157** |

## 4. Conclusions

The evaluation of Bing, ChatGPT, and GPT-4 in responding to GRE test questions, containing quantitative and verbal reasoning questions, has revealed insights into the capability of these AI chatbots. However, GPT-4 demonstrates a well-rounded performance across a range of quantitative and verbal tasks, maintaining a consistently high level of accuracy. It is evident that GPT-4 has demonstrated superior performance, and it outperformed both Bing and ChatGPT in arithmetic, algebra, geometry, and data analysis across all styles. Bing's performance in image-based quantitative questions was relatively better than ChatGPT, which struggled to interpret the external images for many questions. Bing also demonstrated notable stability and the lowest level of uncertainty when requested to reanalyze its results for verification. In verbal reasoning, Bing's performance, particularly in the medium category, was noteworthy, but it faced challenges in the easy category. Conversely, ChatGPT's performance improving as the difficulty level increased. GPT-4's performance indicates its ability to handle a broad range of verbal questions, especially in easy category including straightforward ones that rely on basic language principles. Its proficiency in handling language structures was evident, outperforming Bing and ChatGPT across the board.

However, both Bing and ChatGPT can provide valuable assistance in achieving high score on the GRE exam despite showing intermediate accuracy. The results provide OpenAI and Microsoft with insights into the areas where their chatbots face challenges, help in enhancing their performance in the studied skills. The findings also suggest the use of AI chatbots, particularly GPT-4, in educational settings and standardized test preparation. However, additional research is required to gain insights into the reasons behind the chatbots' inability to solve certain questions, paving the way for further development and improvement of these chatbot systems. Moreover, it is essential to carefully consider the adaptability of online testing, especially in emergency situations like the Covid pandemic, in order to mitigate the risk of cheating incidents among students who heavily depend on these chatbots.

**Funding:** This research received no external funding.

**Conflicts of Interest:** The authors declare no conflict of interest in the design of the study; in the collection, analysis, or interpretation of data; in the writing of the manuscript, or in the decision to publish the results.

## 5. References

Abro, W. A., Aicher, A., Rach, N., Ultes, S., Minker, W., & Qi, G. (2022). Natural language understanding for argumentative dialogue systems in the opinion building domain. *Knowledge-Based Systems*, *242*, 108318. https://doi.org/https://doi.org/10.1016/j.knosys.2022.108318

Ali, M. Y., Naeem, S. Bin, & Bhatti, R. (2020). Artificial intelligence tools and perspectives of university librarians: An overview. *Business Information Review*, *37*(3), 116–124. https://doi.org/https://doi.org/10.1177/0266382120952016

Ali, S. R., Dobbs, T. D., Hutchings, H. A., & Whitaker, I. S. (2023). Using ChatGPT to write patient clinic letters. *The Lancet Digital Health*, *5*(4), e179–e181. https://doi.org/https://doi.org/10.1016/S2589-7500(23)00048-1

Baidoo-Anu, D., & Ansah, L. O. (2023). Education in the era of generative artificial intelligence (AI): Understanding the potential benefits of ChatGPT in promoting teaching and learning. *Journal of AI*, *7*(1), 52–62.

Biswas, S. S. (2023a). Potential use of chat gpt in global warming. *Annals of Biomedical Engineering*, *51*(6), 1126–1127. https://doi.org/https://doi.org/10.1007/s10439-023-03171-8

Biswas, S. S. (2023b). Role of chat gpt in public health. *Annals of Biomedical Engineering*, *51*(5), 868–869. https://doi.org/https://doi.org/10.1007/s10439-023-03172-7

Bleske-Rechek, A., & Browne, K. (2014). Trends in GRE scores and graduate enrollments by gender and ethnicity. *Intelligence*, *46*, 25–34. https://doi.org/https://doi.org/10.1016/j.intell.2014.05.005

Burger, B., Kanbach, D. K., Kraus, S., Breier, M., & Corvello, V. (2023). On the use of AI-based tools like ChatGPT to support management research. *European Journal of Innovation Management*, *26*(7), 233–241. https://doi.org/https://doi.org/10.1108/EJIM-02-2023-0156

Choi, J. H., Hickman, K. E., Monahan, A., & Schwarcz, D. (2023). Chatgpt goes to law school. *Available at SSRN*.

Chowdhary, K., & Chowdhary, K. R. (2020). Natural language processing. *Fundamentals of Artificial Intelligence*, 603–649. https://doi.org/https://doi.org/10.1007/978-81-322-3972-7_19

Crompton, H., & Burke, D. (2023). Artificial intelligence in higher education: the state of the field. *International Journal of Educational Technology in Higher Education*, *20*(1), 1–22. https://doi.org/https://doi.org/10.1186/s41239-023-00392-8

Dominy, C. L., Arvind, V., Tang, J. E., Bellaire, C. P., Pasik, S. D., Kim, J. S., & Cho, S. K. (2022). Scoliosis surgery in social media: a natural language processing approach to analyzing the online patient perspective. *Spine Deformity*, 1–8. https://doi.org/https://doi.org/10.1007/s43390-021-00433-0

ETS. (2023a). *Educational Testing Service - Official Website*. https://www.ets.org/

ETS. (2023b). *GRE® Verbal Reasoning and Quantitative Reasoning Sample Questions with Explanations*. Educational Testing Service. https://www.ets.org/pdfs/gre/gre-sample-questions.pdf

ETS. (2023c). *Official GRE® Quantitative Reasoning Practice Questions Volume 1* (2nd ed.). Educational Testing Service. https://www.ets.org/gre/test-takers/general-test/prepare/prep-books-services.html

ETS. (2023d). *Official GRE® Verbal Reasoning Practice Questions Volume 1* (2nd ed.). Educational Testing Service. https://www.ets.org/gre/test-takers/general-test/prepare/prep-books-services.html

Fang, H., Lu, W., Wu, F., Zhang, Y., Shang, X., Shao, J., & Zhuang, Y. (2015). Topic aspect-oriented summarization via group selection. *Neurocomputing*, *149*, 1613–1619. https://doi.org/https://doi.org/10.1016/j.neucom.2014.08.031

Gilson, A., Safranek, C. W., Huang, T., Socrates, V., Chi, L., Taylor, R. A., & Chartash, D. (2023). How does ChatGPT perform on the United States medical licensing examination? The implications of large language models for medical education and knowledge assessment. *JMIR Medical Education*, *9*(1), e45312. https://doi.org/https://doi.org/10.2196/45312

Google. (2023). *Bard Chatbot [AI Language Model]*. https://bard.google.com/chat

Haman, M., & Školník, M. (2023). Using ChatGPT to conduct a literature review. *Accountability in Research*, 1–3. https://doi.org/https://doi.org/10.1080/08989621.2023.2185514

Hargreaves, S. (2023). 'Words Are Flowing Out Like Endless Rain into a Paper Cup': ChatGPT & Law School Assessments. *The Chinese University of Hong Kong Faculty of Law Research Paper*, *2023–03*.

Hirschberg, J., & Manning, C. D. (2015). Advances in natural language processing. *Science*, *349*(6245), 261–266. https://doi.org/https://doi.org/10.1126/science.aaa8685

Khurana, D., Koli, A., Khatter, K., & Singh, S. (2023). Natural language processing: State of the art, current trends and challenges. *Multimedia Tools and Applications*, *82*(3), 3713–3744. https://doi.org/https://doi.org/10.1007/s11042-022-13428-4

Klieger, D. M., Kotloff, L. J., Belur, V., Schramm-Possinger, M. E., Holtzman, S. L., & Bunde, H. (2022). Studies of Possible Effects of GRE® ScoreSelect® on Subgroup Differences in GRE® General Test Scores. *ETS Research Report Series*, *2022*(1), 1–33. https://doi.org/https://doi.org/10.1002/ets2.12356

Kung, T. H., Cheatham, M., Medenilla, A., Sillos, C., De Leon, L., Elepaño, C., Madriaga, M., Aggabao, R., Diaz-Candido, G., & Maningo, J. (2023). Performance of ChatGPT on USMLE: Potential for AI-assisted medical education using large language models. *PLoS Digital Health*, *2*(2), e0000198. https://doi.org/https://doi.org/10.1371/journal.pdig.0000198

Liu, O. L., Bridgeman, B., Gu, L., Xu, J., & Kong, N. (2015). Investigation of response changes in the GRE revised general test. *Educational and Psychological Measurement*, *75*(6), 1002–1020. https://doi.org/https://doi.org/10.1177/0013164415573988

Moneta-Koehler, L., Brown, A. M., Petrie, K. A., Evans, B. J., & Chalkley, R. (2017). The limitations of the GRE in predicting success in biomedical graduate school. *PLoS One*, *12*(1), e0166742. https://doi.org/https://doi.org/10.1371/journal.pone.0166742

OpenAI. (2023a). *ChatGPT [AI Language Model]*. OpenAI. https://chat.openai.com

OpenAI. (2023b). *GPT-4 [AI Language Model]*. OpenAI. https://openai.com/gpt-4

Ouyang, F., Wu, M., Zheng, L., Zhang, L., & Jiao, P. (2023). Integration of artificial intelligence performance prediction and learning analytics to improve student learning in online engineering course. *International Journal of Educational Technology in Higher Education*, *20*(1), 1–23. https://doi.org/https://doi.org/10.1186/s41239-022-00372-4

Phillips-Wren, G. (2012). AI tools in decision making support systems: a review. *International Journal on Artificial Intelligence Tools*, *21*(02), 1240005. https://doi.org/https://doi.org/10.1142/S0218213012400052

Roohr, K., Olivera-Aguilar, M., Bochenek, J., & Belur, V. (2022). Exploring GRE® and TOEFL® score profiles of international students intending to pursue a graduate degree in the United States. *ETS Research Report Series*, *2022*(1), 1–27. https://doi.org/https://doi.org/10.1002/ets2.12343

Rudolph, J., Tan, S., & Tan, S. (2023). War of the chatbots: Bard, Bing Chat, ChatGPT, Ernie and beyond. The new AI gold rush and its impact on higher education. *Journal of Applied Learning and Teaching*, *6*(1). https://doi.org/https://doi.org/10.37074/jalt.2023.6.1.23

Surameery, N. M. S., & Shakor, M. Y. (2023). Use chat gpt to solve programming bugs. *International Journal of Information Technology & Computer Engineering (IJITC) ISSN: 2455-5290*, *3*(01), 17–22. https://doi.org/https://doi.org/10.55529/ijitc.31.17.22

Susnjak, T. (2022). ChatGPT: The end of online exam integrity? *ArXiv Preprint ArXiv:2212.09292*. https://doi.org/https://doi.org/10.48550/arXiv.2212.09292